\documentclass[preprint,nopreprintline,12pt]{elsarticle}

\usepackage{amssymb}
\usepackage{amsmath}
\usepackage{booktabs}
\usepackage{todonotes}
\usepackage{hyperref}
\usepackage{xcolor}
\hypersetup{
  colorlinks   = true, 
  urlcolor     = red, 
  linkcolor    = blue, 
  citecolor   = blue 
}
\usepackage{subcaption}

\journal{Nuclear Physics B}

\begin{document}

\begin{frontmatter}



\title{PARASIDE: An Automatic Paranasal Sinus Segmentation and Structure Analysis Tool for MRI}
 
\author[mri,aimed]{Hendrik Möller}
\affiliation[mri]{organization={Department for Interventional and Diagnostic Neuroradiology, TUM University Hospital},
             addressline={Ismaninger Straße 22},
             city={Munich},
             postcode={81675},
             state={Bavaria},
             country={Germany}}

\affiliation[aimed]{organization={Chair for AI in Healthcare and Medicine},
            addressline={Technical University of Munich (TUM) and TUM University Hospital}, 
            city={Munich},
            postcode={81675}, 
            state={Bavaria},
            country={Germany}}

\affiliation[gr]{organization={Department of Otorhinolaryngology, Head and Neck Surgery, University Medicine Greifswald}, 
             addressline={Ferdinand-Sauerbruch-Strasse},
             city={Greifswald},
             postcode={17475},
             state={Mecklenburg-Western Pomerania},
             country={Germany}}
             
\affiliation[mue]{organization={Department of Otorhinolaryngology, Head and Neck Surgery, University Hospital Muenster},
             addressline={Kardinal-von-Galen-Ring 10},
             city={Muenster},
             postcode={48149},
             state={North Rhine-Westphalia},
             country={Germany}}             

\affiliation[bigr]{organization={Institute of Bioinformatics},
             addressline={Felix-Hausdorff-Str. 8},
             city={Greifswald},
             postcode={17475},
             state={Mecklenburg-Western Pomerania},
             country={Germany}}
\affiliation[zurich]{organization={Department of Quantitative Biomedicine, University of Zurich},
             addressline={Winterthurerstrasse 190},
             city={Zurich},
             postcode={8057},
             state={},
             country={Switzerland}}
\affiliation[imperial]{organization={Department of Computing, Imperial College London},
             addressline={180 Queen's Gate},
             city={London},
             postcode={SW7 2AZ},
             state={},
             country={England}}

\author[gr]{Lukas Krautschick} 
\author[mri,aimed]{Matan Atad} 
\author[mri,aimed]{Robert Graf} 
\author[gr]{Chia-Jung Busch} 
\author[gr,mue]{Achim Beule} 
\author[gr]{Christian Scharf} 
\author[bigr]{Lars Kaderali} 
\author[zurich]{Bjoern Menze} 
\author[aimed,imperial]{Daniel Rueckert} 
\author[mri]{Jan Kirschke} 
\author[gr]{Fabian Schwitzing} 

\begin{abstract}
Chronic rhinosinusitis (CRS) is a common and persistent sinus imflammation that affects 5 - 12\% of the general population. It significantly impacts quality of life and is often difficult to assess due to its subjective nature in clinical evaluation. We introduce PARASIDE, an automatic tool for segmenting air and soft tissue volumes of the structures of the sinus maxillaris, frontalis, sphenodalis and ethmoidalis in T1 MRI. By utilizing that segmentation, we can quantify feature relations that have been observed only manually and subjectively before. We performed an exemplary study and showed both volume and intensity relations between structures and radiology reports. While the soft tissue segmentation is good, the automated annotations of the air volumes are excellent. The average intensity over air structures are consistently below those of the soft tissues, close to perfect separability. Healthy subjects exhibit lower soft tissue volumes and lower intensities. Our developed system is the first automated whole nasal segmentation of 16 structures, and capable of calculating medical relevant features such as the Lund-Mackay score.
\end{abstract}


\end{frontmatter}


\section{Introduction}
\label{introduction}


Magnetic resonance imaging (MRI) is a vital diagnostic tool, offering superior soft tissue contrast without ionizing radiation, which is particularly beneficial for children and patients with cystic fibrosis \cite{Pfaar.2023_AWMFS2K}. Computer tomography (CT) and MRI provide complementary information in sinonasal evaluations \cite{MOSSABASHA201314_NNH_imaging}. CT excels at depicting osseous involvement, erosion, and surgical mapping, while MRI differentiates between inflammatory disease and sinonasal masses, illustrating the boundaries and extent of a tumor for conditions such as chronic rhinosinusitis (CRS).

Segmentation of paranasal sinuses and pathologies improves disease evaluation \cite{Cellina.2021_review_segmentation} and supports the development of objective scoring systems essential for personalized medicine and individualized treatments, such as biologics \cite{Fokkens.2020_EPOS2020}. However, these scores are often limited to research settings and lack clinical integration. In big data studies like the "Study of Health in Pomerania" (SHIP) \cite{Volzke.2022_SHIP_start_trend}, automatic segmentation is essential to streamline analysis and integrate large-scale datasets. Advanced imaging and segmentation technologies support precise monitoring, stratification, and personalized therapies, addressing the demands of data-driven healthcare.

The SHIP includes two main cohorts, SHIP-START (1997–2001) and SHIP-TREND (2008–2012)\cite{Volzke.2022_SHIP_start_trend}, aimed at monitoring the health of the adult population in northeast Germany. The study evaluates the prevalence and incidence of risk factors, subclinical and clinical diseases, as well as their interrelationships. A third cohort, SHIP-NEXT, was launched in 2021 with 4,000 participants aged 20 to 79 years \cite{Volzke.2021_SHIP_next}. However, SHIP-NEXT is not included in this paper as data collection has only recently begun.

Deep Learning techniques are commonly used to address the problem of automatic semantic segmentation \cite{segmentation_survey}. Though so far, most existing models for nasal segmentation only work on cone beam computed tomographic (CBCT) images \cite{bui2015nose_ct_seg, whangbo_strong_ct_seg} or focus only on one structure \cite{huang_ct_seg, ozturk2024_ct_maxillary_sinus_seg}. Furthermore, there haven't been many attempts to automatically segment and analyze pathologies in the nasal cavities \cite{altun2024maxillary_pathology_seg}. 
However, an automated approach to paranasal sinus segmentation is essential to overcome the labor-intensive nature of manual methods \cite{Brzoska.2022_Parotis_segmentation, Caspar.2015_manual_segmentation_KH, Schneider.2018_manual_segmentation_KH_STH}. These studies highlight the need for efficient and scalable solutions to improve consistency and enable large-scale data analysis.

The purposes of this study are (1), to establish PARASIDE, the first automated Paranasal Sinus segmentation approach with 16 different structures for MRI, (2) to show that the segmentation approach can directly detect and localize pathological tissue, (3), to demonstrate how this segmentation enables downstream analysis and derive important medical characteristics, and (4) to make both the segmentation masks and the implementation publicly available for others to use.


\section{Materials and methods}
\label{methods}


\subsection{Data}

The data for this study were obtained from the SHIP-START and SHIP-TREND cohorts of SHIP, a population-based study in Northeast Germany. From the total cohort of 8,728 participants with 5600 T1 weighted MRI images, we focused on a balanced random subset of 273 subjects, evenly distributed across the four subcohorts and defined by a script to ensure balance in sex, age, and the presence of pathologies. SHIP is conducted in accordance with the Declaration of Helsinki. The ethics committee of the University of Greifswald approved the study protocol. The T1-weighted images of the head were acquired using the sequence Kopf\_T1\_mpr\_tra\_iso\_p2. This protocol utilized a repetition time (TR) of 1900 ms and an echo time (TE) of 3:37 ms, with a slice thickness (SD) of 1 mm. The total scan duration for this sequence was approximately 3 minutes and 38 seconds.


In addition to the imaging data, we obtained information on the diagnosis of chronic rhinosinusitis in the maxillary and frontal sinus as a radiology report. Further details on study design, recruitment, and data collection are provided in Völzke et al. \cite{Volzke.2022_SHIP_start_trend}. The SHIP data committee approved our application for data usage, after which we received the raw data, including MRI images in DICOM format. These images were subsequently converted into NIFTI format.

We created a balanced training set of 100 participants and a test set of 60 participants, ensuring diversity in demographics and pathologies to optimize model performance and generalization (see Table \ref{tab:demo}). The majority of our subset of data contains pathologies. The remaining 113, together with the test set, make up our analysis set for downstream experiments.

Manual annotations for the dataset were created through the collaborative efforts of two experts (**blinded initials**), with one providing oversight and guidance based on 7 years of experience while the other contributed under close supervision to ensure accuracy and consistency. They used ITKSnap as an annotation tool \cite{yushkevich2006itksnap}. Notably, the experts did not consult the SHIP reports for their annotation and thus annotated pathologies by assessing the subjects themselves. 


These annotations included 4 main structures: Sinus maxillaris, frontalis, sphenoidalis, and ethmoidalis. We split each of them into left and right with additional separation of air and soft tissue (including pathological tissue), resulting in a total of 16 distinct annotated structures.

\subsection{Segmentation Approach}

\begin{figure}[htbp]
    \centering
    \includegraphics[width=0.8\textwidth]{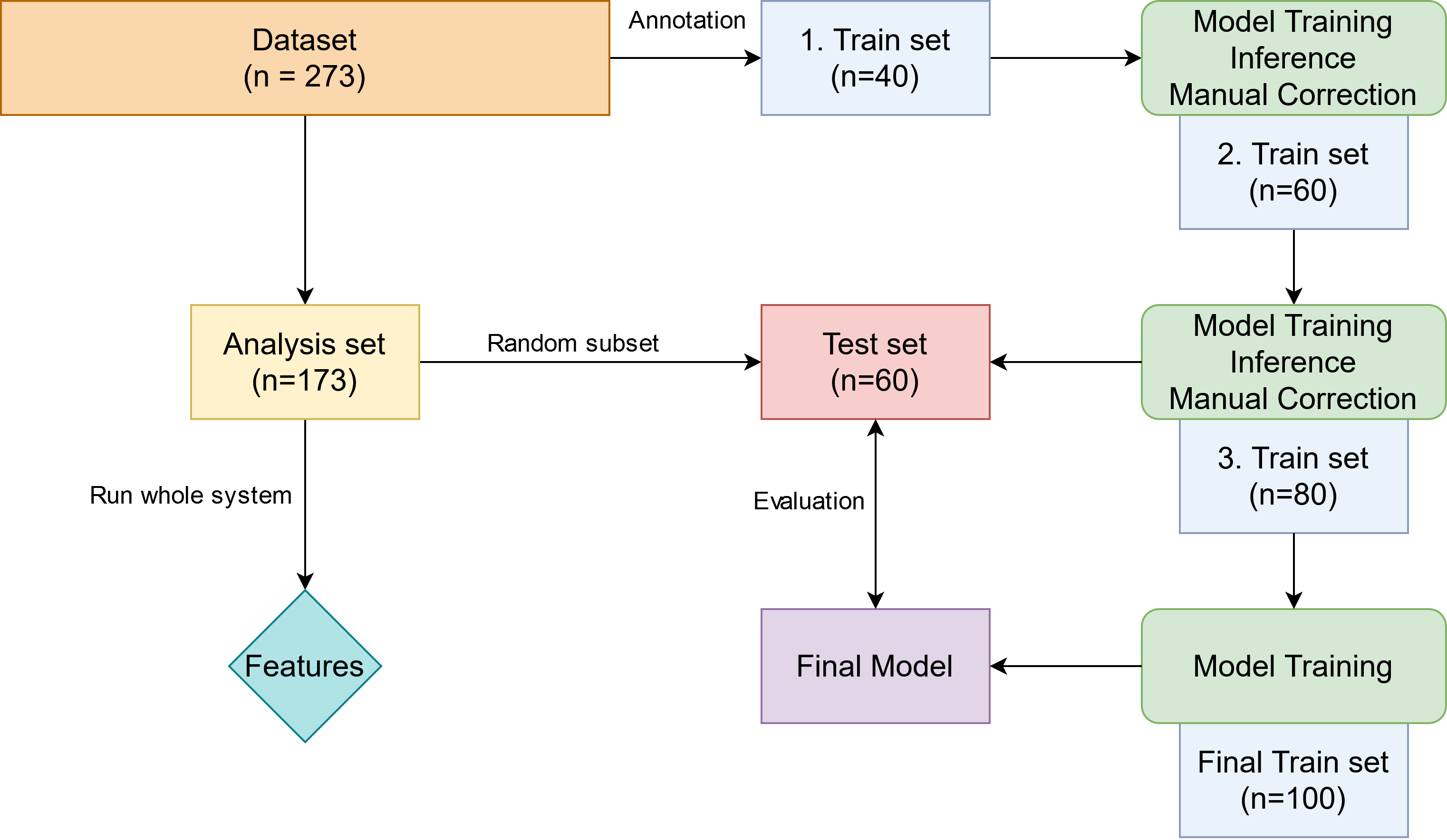}
    \caption{The flow of our utilized data. We only use one dataset and an iterative approach for data annotation and quality control. The final model is evaluated against the test split. n stands for number of subjects.}
    \label{fig:dataflow}
\end{figure}

We adopted an iterative annotation and training process. Initially, our two experts annotated 20 subjects each. With a total of 40 training subjects, we trained a nnUNet \cite{isensee2021nnunet} and used that model on our training set. We then corrected these masks and reviewed the largest discrepancies for already annotated subjects to increase our data quality and quantity (see Figure \ref{fig:dataflow} for our iteration process).

For training the segmentation model, we used the parameters proposed by nnUNet other than setting the patch size to $(128,128,128)$. No data preprocessing was necessary, as the samples already had a consistent volume size of $(256,176,176)$ and resolution $(1,1,1)$. However, we multiplied our training data tenfold by using horizontal flips and elastic deformation, which were computed prior to training. When horizontal flipping, we flip the corresponding left and right labels of structures to avoid erroneous training data. We trained for 5000 epochs.

Once our experts had annotated all 100 training samples and did not find any mistakes anymore, we trained one final time, using the whole training split.

\subsection{Downstream Utility}

Based on our segmentation, we can compute medical-relevant scores. For an exemplary study, we decided to calculate the following features for each structure: Volume of each structure, intensity average and standard deviation for each structure, and bounding box measurements. 
From these features, we can directly calculate medical labels such as hypoplasia (below 5\% of normal volume), aplasia (absence of structure), and the widespread Lund-Mackay staging system \cite{Lund.1997_LundMackayScore}. The Lund-Mackay score is a subjective staging system typically assigned by a radiologist in CT scans \cite{Hopkins2007_Interpretation-LMS}, has also been validated for use with MRI, demonstrating comparable diagnostic accuracy \cite{Lin_LMS_in_MRI}. Our approach aims to make this scoring system objective. We retain the same scoring levels of 0/1/2 but define them quantitatively as follows: less than 5\% opacification (0), between 5\% and 95\% opacification (1), and above 95\% opacification (2). Unlike the traditional system, we do not separate the anterior ethmoid, posterior ethmoid, and osteomeatal complex. To compare our total sum score with the traditional Lund-Mackay score, we multiplied our ethmoidal Lund-Mackay score by 3 to achieve a comparable overall score (Modified Lund-Mackay).

\subsection{Experiments}

We used our final trained model on the test data of 60 subjects and evaluated the segmentation performance using the Dice similarity coefficient (DSC) and the average symmetric surface distance (ASSD). To that end, we utilize panoptica \cite{kofler2023panoptica}, a tool that calculates all those metrics and more. We calculate the metrics for each of the structures individually, then average together the left- and right variants of the same structure. 

We do not provide competing algorithms. To the best of our knowledge, there is no publicly available algorithm to compare against. 

Additionally, we ran the model on the 173 subjects of our analysis set and calculated the features using our predicted segmentation. We then report mean values of features and provide insights. When analyzing the data, we always treat the left and right variants of the same structure as two individual data points, thus setting our data size to 346. We compare the volumes and intensity averages of the maxillaris and frontalis structure to the SHIP radiology reports. We excluded 3 subjects, as they were marked in the reports as "technical", meaning the radiologist could not assess it.



\section{Results and Discussion}
\label{results}


\subsection{Segmentation Performance}
Our segmentation model trained on 100 manually annotated subjects has an average DSC of 0.95 over air volume across all eight structures on our test data (see Table \ref{tab:segmentation}). On the other hand, the soft tissues only reach a DSC of 0.56. Additionally, while the values are all above 0.9 for the air volume, with consistently low standard deviations, the soft tissue segmentation shows larger differences between structures. The Air tissue of the sinus maxillaris region performs best, while the soft tissue (ST) of the sphenoidalis performs the worst.

\begin{figure}[htbp]
    \centering
    \includegraphics[width=1.0\textwidth]{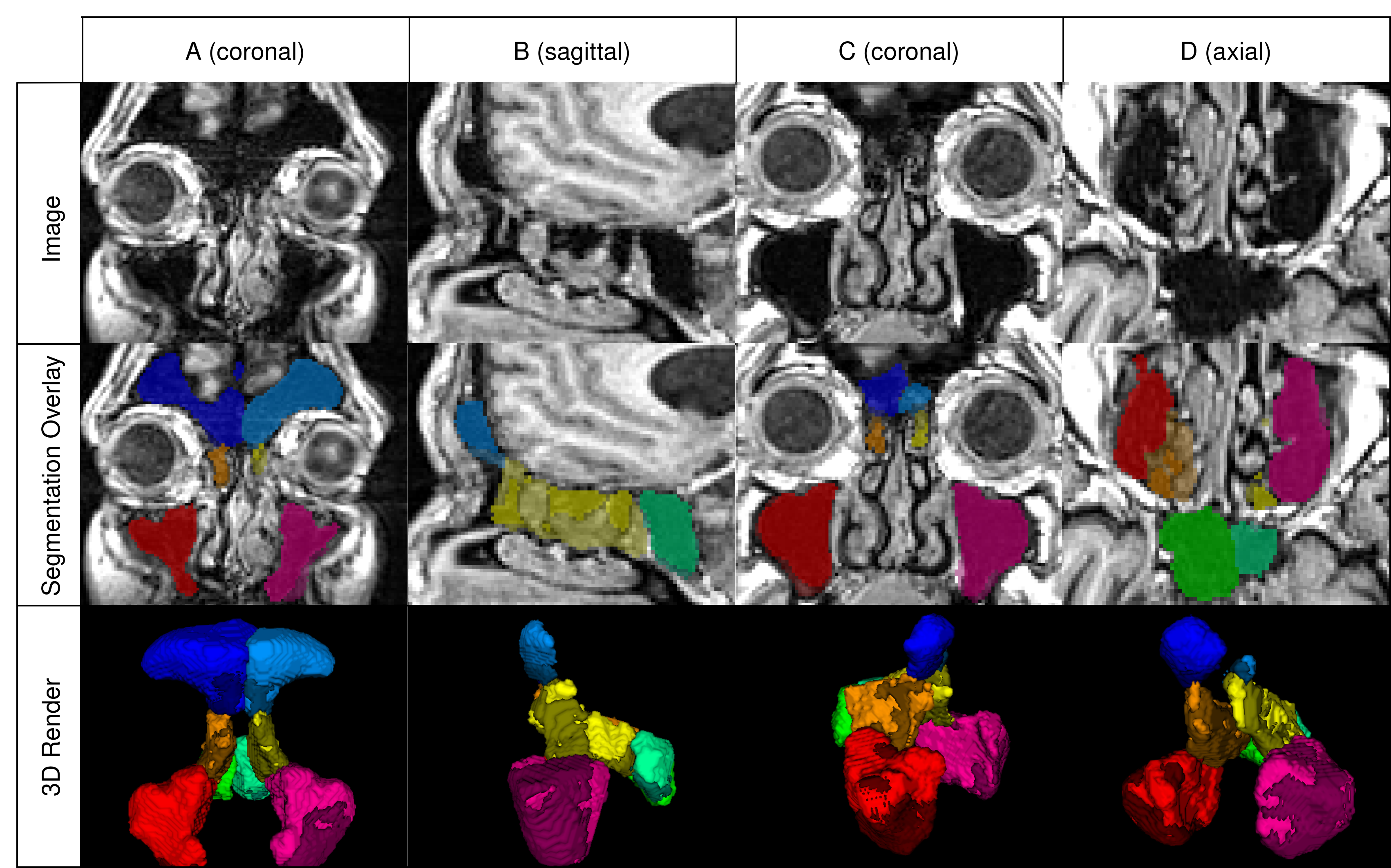}
    \caption{Showcase of our predicted segmentation masks for random samples from the test set (A -- D). While A and C show a coronal view, B is a sagittal view, and D is an axial one. The rows are, in order, the image, then the image overlayed with the predicted annotation of our final mode, and a 3D view of the segmentation. The colors are the different structures. The air labels are of light colors while the soft tissue ones have the darker shades. Color coding of paranasal sinuses: Red (right maxillary), pink (left maxillary), dark blue (right frontal), light blue (left frontal), orange (right ethmoidal), yellow (left ethmoidal), green (right sphenoidal), turquoise (left sphenoidal).}
    \label{fig:eval-segmentation}
\end{figure}

For some qualitative examples, see Figure \ref{fig:eval-segmentation}.

\subsection{Feature Analysis}

\begin{figure}[!htbp]%
    \centering
    \begin{subfigure}{0.5\textwidth}%
    \includegraphics[width=0.95\linewidth]{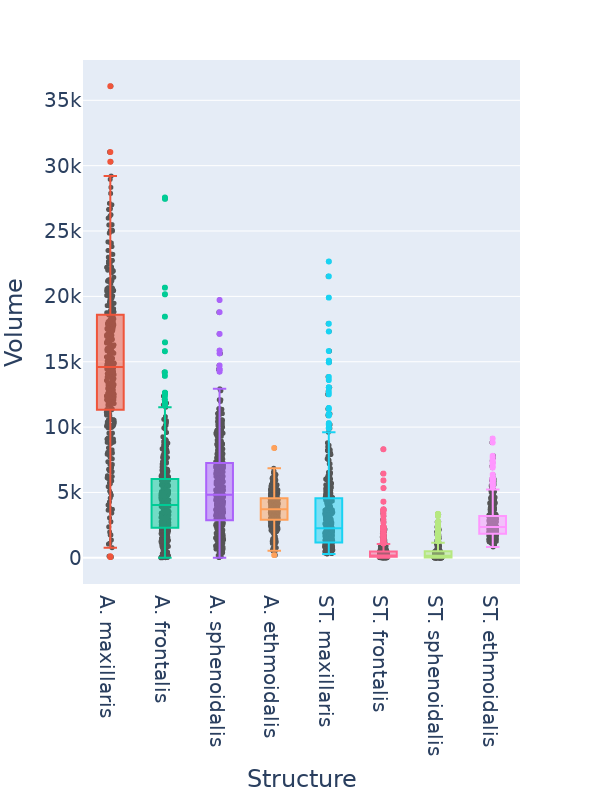}
    \label{fig:eval-boxplotvolume}
    \end{subfigure}%
    \begin{subfigure}{0.5\textwidth}%
    \includegraphics[width=0.95\linewidth]{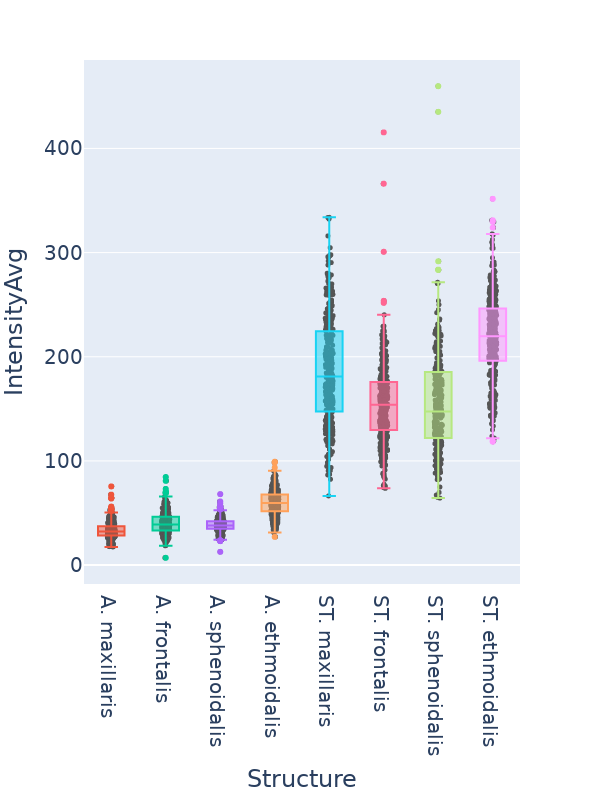}
    \label{fig:eval-boxplotintensity}
    \end{subfigure}%
    \caption{The volume (left) in $mm^3$ and intensity average (right) of different structures across our data. "A." for the aerated/ air-filled space and "ST." for the soft tissue/ pathology proportion. }
    \label{fig:eval-boxplot}
\end{figure}

When analyzing the calculated features, we observe that the average intensity of soft tissue is a lot higher compare to air tissue (see Figure \ref{fig:eval-boxplot}). Even including the standard deviation, we observe a nearly perfect separability between individual structure's air and soft tissue annotations.
The volumes of the A. maxillaris have the largest standard deviations (see Figure \ref{fig:eval-boxplot}), while the soft tissues exhibit mostly smaller volumes. 

\begin{figure}[!htbp]%
    \centering
    \begin{subfigure}{0.5\textwidth}%
    \includegraphics[width=0.95\linewidth]{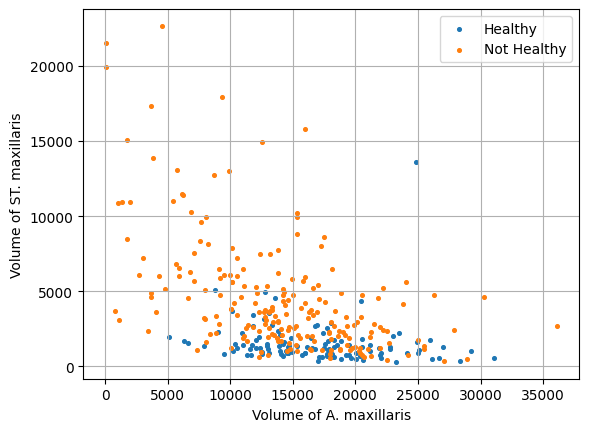}
    \label{fig:eval-volume-maxillaris}
    \end{subfigure}%
    \begin{subfigure}{0.5\textwidth}%
    \includegraphics[width=0.95\linewidth]{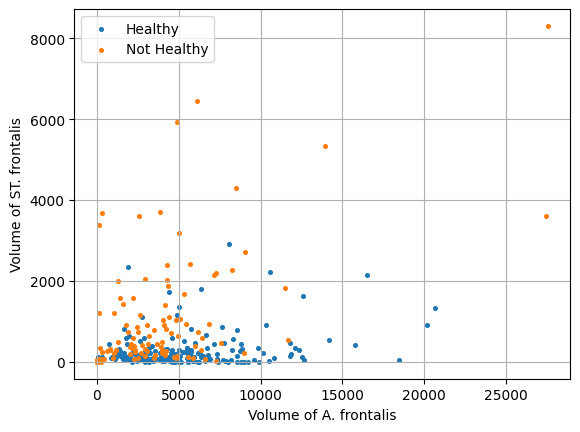}
    \label{fig:eval-volume-frontalis}
    \end{subfigure}%
    \caption{Scatter plots illustrating the relationship between the volume a of A. maxillaris/ frontalis (air-filled or aerated) and ST. maxillaris/ frontalis (soft tissue/pathology). Data points are categorized as "Healthy" (blue) and "Not Healthy" (orange) according to the SHIP radiology reports.}
    \label{fig:eval-volume}
\end{figure}

The scatter plots in Figure \ref{fig:eval-volume} illustrate the relationship between the volume of air-filled sinuses (A. maxillaris and A. frontalis) and their respective soft tissue components (ST. maxillaris and ST. frontalis). The data demonstrate that "Healthy" individuals (blue) generally exhibit lower volumes of soft tissue (ST), whereas "Not Healthy" individuals (orange) show a wider distribution with notably higher ST volumes. This trend is more pronounced in the maxillaris sinus compared to the frontalis sinus. These findings suggest a clear distinction between healthy and pathological cases based on sinus volume metrics.

\begin{figure}[!htbp]%
    \centering
    \begin{subfigure}{0.5\textwidth}%
    \includegraphics[width=0.95\linewidth]{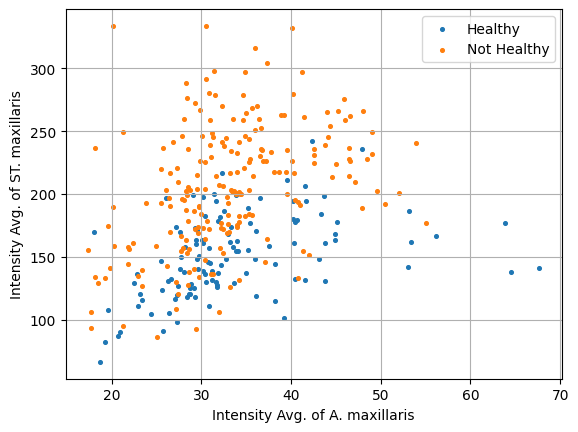}
    \label{fig:eval-intensity-maxillaris}
    \end{subfigure}%
    \begin{subfigure}{0.5\textwidth}%
    \includegraphics[width=0.95\linewidth]{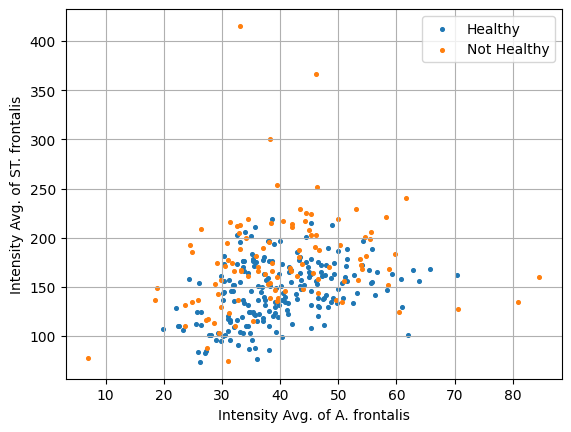}
    \label{fig:eval-intensity-frontalis}
    \end{subfigure}%
    \caption{Scatter plots illustrating the relationship between the average intensity a of A. maxillaris/ frontalis (air-filled or aerated) and ST. maxillaris/ frontalis (soft tissue/pathology). Data points are categorized as "Healthy" (blue) and "Not Healthy" (orange) according to the SHIP radiology reports.}
    \label{fig:eval-intensity}
\end{figure}

Figure \ref{fig:eval-intensity} illustrates the relationship between the average intensity of air-filled (A. maxillaris and A. frontalis) and soft tissue (ST. maxillaris and ST. frontalis). The scatter plots demonstrate that "Healthy" individuals are generally clustered at lower intensity values for both air-filled and soft tissue components, while unhealthy cases show higher intensity averages, reflecting increased soft tissue presence or denser pathological changes.

\begin{figure}[!htbp]
    \centering
    \includegraphics[width=0.7\textwidth]{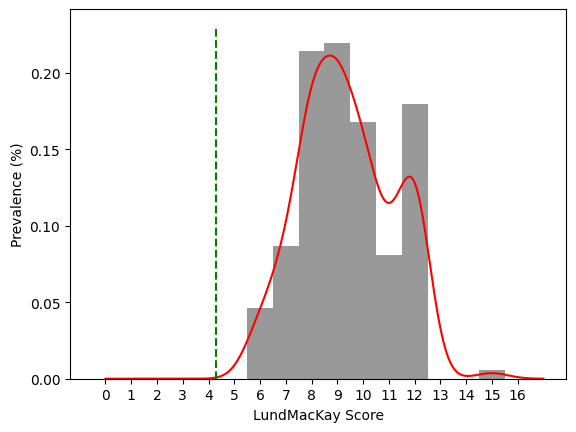}
    \caption{The distribution of our modified Lund-Mackay score calculated over our analysis set using our predicted segmentations. We report the percentage prevalence (gray) and the kernel density estimate (red). The green dashed line corresponds to the mean “normal” Lund-Mackay score (4.3) observed in patients undergoing imaging for nonrhinologic symptoms, as identified in previous studies \cite{Hopkins2007_Interpretation-LMS}.}
    \label{fig:eval-lmc}
\end{figure}

After calculating our modified Lund-Mackay score over our analysis set, we observe a cluster between scores of 7–12, a distribution peak around the score 9 and very few cases at the extreme high end, e.g., 15 or 16 (see Figure \ref{fig:eval-lmc}). Higher LMS indicates greater sinus involvement, likely associated with severe disease. Most individuals fall within moderate disease severity, with few extreme cases of either mild or severe disease.

\subsection{Discussion}

We developed a tool for automated segmentation of paranasal sinuses in T1-weighted MRI images using a U-Net, trained on 100 training and 60 test images. The model demonstrated high accuracy, particularly in segmenting air-filled sinus regions (e.g. mean DICE and mean ASSR 1-8) and soft tissue structures (e.g. mean DICE and mean ASSR Label 9-16). This tool enables precise volume measurement, essential for diagnosing conditions like aplasia (absent sinus) or hypoplasia (underdeveloped sinus with less than 5\% of the average volume). It also provides absolute and relative volume and average intensity analysis to calculate a relation score between air and soft tissue (seen in Table \ref{tab:evalfeatures}). 

The results highlight distinctions between health states based on intensity (see Figure \ref{fig:eval-intensity}) and volume (see Figure \ref{fig:eval-volume}) values of the air-filled and soft tissue components of the sinuses and their complementary roles in distinguishing between healthy and diseased states. Higher soft tissue (ST) volumes and intensities observed in "Not Healthy" cases, such as those with sinusitis or polyposis, reflect pathological changes like inflammation, fluid accumulation, or polyp formation. In the maxillaris sinus, these differences are more pronounced, suggesting its greater sensitivity to disease-related changes compared to the frontalis sinus. The clustering of "Healthy" cases at lower intensities and volumes confirms the absence of significant pathology, while the overlap in certain cases indicates the need for refined thresholds or additional features. These findings further validate the utility of intensity-based metrics in identifying pathological states while underscoring the need to address potential technical and anatomical confounders.

In addition, similar patterns were observed in our modified Lund-Mackay Score analysis (Figure \ref{fig:eval-lmc}). We modified the Lund-Mackay staging system into an objective framework using automated segmentation, defining scores quantitatively ($<$5\% = 0; 5–95\% = 1; $>$95\% = 2). Due to our manual segmentation approach, which did not separate anterior/posterior ethmoid or the osteomeatal complex, we simplified the staging process while maintaining clinical relevance. To ensure comparability, we adjusted our total score by multiplying the ethmoidal score by 3. This method enhances consistency and objectivity in staging chronic rhinosinusitis. The prevalence distribution as shown in Figure \ref{fig:eval-lmc}) suggests that the study population mostly experiences moderate sinus involvement based on LMS, with severe pathology and no disease activity being less common. Interestingly, our findings show no cases with a completely healthy LMS, despite including healthy MRI subjects. This discrepancy could be due to multiplying the ethmoidal score by 3, which may overestimate the LMS intensity and the high volume of ST. ethmoidalis observed in our data (seen in Table \ref{tab:evalfeatures}), likely influenced by the absence of segmentation into three ethmoidal subparts. Alternatively, it may indicate the need to adjust the current LMS thresholds ($<$5 / 5–95 / $>$95) to better capture healthy cases. This reflects the composition of our training dataset, as we included a higher proportion of pathological findings and subjects to optimize the detection of pathologies.

Furthermore the tool counts distinct mucosal structures to evaluate abnormalities, offers size and shape metrics including height, width, and depth for structural assessments, and calculates spatial relationships between sinus structures to detect patterns in localized pathologies. This development significantly enhances sinus imaging, offering precise diagnostics and advancing research for individualized treatment approaches.


In conclusion, while most studies on paranasal sinus segmentation have focused on CT imaging due to its superior visualization of osseous structures \cite{Tingelhoff.2008_manual_segmentation_approaches, Iwamoto.2019_probalistic_atlas_FCN}, there is a notable gap in comparable work using T1-weighted MRI. Our study addresses this void by demonstrating the potential of MRI as a radiation-free alternative, particularly suited for longitudinal studies, as highlighted by Andersen et al. \cite{Andersen.2018_comparison_MRI_CT}. Previous research has primarily relied on manual segmentation \cite{Andersen.2018_comparison_MRI_CT, Tingelhoff.2008_manual_segmentation_approaches}, which, while accurate, is time-consuming and prone to inter- and intraobserver variability.

Unlike CT-based automated approaches, such as those by Morgan et al. \cite{Morgan.2022_automated_segmentation_CBCT} and Iwamoto et al. \cite{Iwamoto.2019_probalistic_atlas_FCN}, which utilized CNNs and FCNs, our work represents a pioneering effort in applying a U-Net architecture to MRI for both segmentation and disease recognition. This innovation not only enhances diagnostic precision but also introduces a scalable solution for large datasets, such as those in the SHIP study. Given the lack of similar work in MRI segmentation of the paranasal sinuses, our study fills a critical gap in the literature and provides a framework for advancing both research and clinical practice in this field.

Our results indicate a higher prevalence of moderate to severe Lund-Mackay Scores compared to the normal score reported by Emmanuel et al. (0.803 ± 2.90) \cite{Emmanuel2018_Normal-LMS-score} and align with Hopkins et al.'s classification \cite{Hopkins2007_Interpretation-LMS}, reflecting a dataset enriched with pathological cases for training purposes.


Our study has several limitations that should be acknowledged. First, although the dataset is comprehensive, it represents the largest available dataset that includes all paranasal sinuses and surrounding soft tissues, providing a unique resource for segmentation studies. However, the sample size may still not fully capture the variability in paranasal sinus anatomy and pathologies, particularly in rare conditions. Additionally, while the imaging data are limited to T1-weighted MRI, which lacks the resolution of CT scans for assessing bony structures, it remains a valuable resource due to its non-invasive nature. T2-weighted fat-saturated images would provide better soft tissue contrast in MRI, and CT scans still hold the gold standard for evaluating bony anatomy. However, the SHIP dataset, with its T1-weighted images and absence of radiation, offers the only standardized imaging option without associated risks.
The dataset was derived from a population in Northeast Germany, which may limit the generalizability of our findings to populations with differing demographics or healthcare settings. 

Certain clinical factors, such as treatment history, environmental exposures, and comorbidities, were not included in this analysis and may influence the model's performance. Future studies should incorporate these factors to enhance the clinical applicability and reliability of our findings.


The clinical relevance of our study lies in its potential to improve the diagnosis and management of paranasal sinus conditions, particularly chronic rhinosinusitis. By utilizing AI-based segmentation of T1-weighted MRI data, we enhance diagnostic accuracy, streamline workflows, and support personalized treatment. We aim to transfer our methods to CT scans and integrate them with clinical data for a more comprehensive assessment. Additionally, we are developing a new score that combines clinical and imaging data to provide actionable recommendations. These findings will extend beyond the paranasal sinuses to the upper airway and other ENT areas, such as the parotid glands, further enhancing patient care.

\subsection{Conclusion}

In conclusion, we developed PARASIDE, a U-Net-based tool for automated segmentation of paranasal sinuses in T1-weighted MRI images, achieving high accuracy in both air-filled and soft tissue regions. The tool provides valuable diagnostic metrics, including volume assessment, opacification scores based on the modified Lund-Mackay scoring system, and structural and spatial analyses. By streamlining sinus imaging and enabling precise, reproducible evaluations, this tool represents a significant step forward in personalized diagnostics and research, paving the way for improved treatment strategies for sinus-related conditions.
Finally, we have published the model weights and the manual annotations, thus making PARASIDE easy to use and advance by others in the future (**blinded hyperlink**).
\section{Tables with captions}
\label{tables}

\begin{table}[hbpt]
    \centering
    \footnotesize
    \begin{tabular}{lccc}
        \toprule
        & SHIP subset                   & train set & test set \\
        \midrule
        Subjects                        & 273 & 100 & 60\\
        Sex (\% female)                 & 65 & 64 & 64\\
        Age range (yrs)                 & 11 -- 88 & 11 -- 88 & 28 -- 83\\
        Mean age (yrs) $\pm$ SD         & $55 \pm 14$ & $56 \pm 14$ & $55 \pm 14$\\
        Pathology Rate (\% has pathology) & 73 & 79 & 73 \\
        \bottomrule
    \end{tabular}
    \caption{The data demographics of this study. We use a subset of the SHIP data which we split into a train and test split. The data not used in the splits combined with the test set are used in a downstream experiment.}
    \label{tab:demo}
\end{table}

\begin{table}[hbpt]
    \centering
    \footnotesize
    \begin{tabular}{lcc}
        \toprule
        Structure & DSC $\uparrow$ & ASSD $\downarrow$ \\
        \midrule
        A. maxillaris           & $0.98 \pm 0.02$    & $0.12 \pm 0.06$ \\
        A. frontalis            & $0.93 \pm 0.10$    & $0.33 \pm 0.27$ \\
        A. sphenoidalis         & $0.96 \pm 0.07$    & $0.27 \pm 0.49$ \\
        A. ethmoidalis          & $0.93 \pm 0.04$    & $0.29 \pm 0.13$ \\
        \midrule
        ST. maxillaris           & $0.74 \pm 0.20$    & $1.07 \pm 1.34$ \\
        ST. frontalis            & $0.50 \pm 0.28$    & $1.24 \pm 1.13$ \\
        ST. sphenoidalis         & $0.25 \pm 0.31$    & $1.99 \pm 1.90$ \\
        ST. ethmoidalis          & $0.78 \pm 0.08$    & $0.60 \pm 0.30$ \\
        \midrule
        Average (Air)          & $0.95 \pm 0.07$    & $0.25 \pm 0.30$ \\
        Average (Soft Tissue)  & $0.56 \pm 0.32$    & $1.12 \pm 1.28$ \\
        \midrule
        Average (overall)       & $0.76 \pm 0.30$    & $0.65 \pm 0.99$ \\
        \bottomrule
    \end{tabular}
    \caption{Performance of our segmentation model for paranasal sinus structures. Metrics include Dice Similarity Coefficient (DSC, higher is better, $\uparrow$) and Average Symmetric Surface Distance (ASSD, lower is better, $\downarrow$). Mean and standard deviation are reported for each structure, with left and right labels evaluated independently and averaged for structure-wise metrics. Results are presented for air (A.) and soft tissue (ST.) structures, as well as overall averages.}
    \label{tab:segmentation}
\end{table}

\begin{table}[hbpt]
    \centering
    \footnotesize
    \begin{tabular}{lrrrrr} 
        \toprule
        Structure & Intensity & Volume & Depth & Width & Height \\
        && $[mm^3]$ & $[mm]$ & $[mm]$ & $[mm]$ \\
        \midrule
        A. maxillaris           & $33.7$    & 14845 & 39.0 & 30.5 & 38.8\\
        A. frontalis            & $40.4$    & 4707 & 17.5 & 29.7 & 26.4 \\
        A. sphenoidalis         & $38.5$    & 5439 & 26.8 & 23.2 & 27.1 \\
        A. ethmoidalis          & $59.8$    & 3735 & 41.4 & 17.7 & 29.2 \\
        \midrule
        ST. maxillaris           & $186.7$    & 3474 & 33.5 & 29.3 & 38.5 \\
        ST. frontalis            & $155.0$    & 530 & 11.8 & 15.3 & 15.1 \\
        ST. sphenoidalis         & $157.6$    & 482 & 14.3 & 13.2 & 13.3\\
        ST. ethmoidalis          & $219.2$    & 2760 & 36.8 & 16.5 & 28.7 \\
        \midrule
        Union of Air Structures        & $38.7$    & 57316 & 74.1 & 83.9 & 87.1 \\
        Union of Soft Tissue Structures  & $203.1$    & 14219.3 & 61.0 & 84.8 & 77.4\\
        \midrule
        Union of all Structures         & $76.0$    & 71535 & 74.4 & 86.3 & 89.2 \\
        \bottomrule
    \end{tabular}
    \caption{Quantitative analysis of air (A.) and soft tissue (ST.) volumes across different paranasal sinus structures (maxillaris, frontalis, sphenoidalis, ethmoidalis) and their unions, considering only existing structures (i.e., excluding cases of aplasia). Measurements include average intensity, volume (mm³), depth (mm), width (mm) and height (mm). The union of all structures represents the comprehensive volumetric assessment of air and soft tissue components.}
    \label{tab:evalfeatures}
\end{table}

\appendix
\section{Appendix}
\label{app}



\bibliographystyle{elsarticle-num} 
\bibliography{bib}






\end{document}